\documentclass[letterpaper, 10 pt, conference]{ieeeconf}

\IEEEoverridecommandlockouts

\usepackage[hidelinks]{hyperref}
\usepackage{subcaption}
\usepackage{booktabs}       
\usepackage{placeins}
\usepackage[nocompress]{cite}
\usepackage{graphicx}
\usepackage[acronym,shortcuts]{glossaries}
\usepackage{comment}
\usepackage{enumerate}
\usepackage[capitalise]{cleveref}
\usepackage[table,xcdraw,dvipsnames]{xcolor}
\usepackage{siunitx}
\usepackage{bm} 
\usepackage{amsmath,amssymb,amsfonts}
\usepackage[english]{babel}
\usepackage{multirow}
\usepackage{pifont}
\usepackage{algorithm}
\usepackage{algpseudocode}
\usepackage{cuted}
\usepackage{cases}
\usepackage{textcomp}
\usepackage{xcolor}
\usepackage{fancyhdr}
\usepackage{tikz}
\usepackage{eso-pic}
\usepackage{dblfloatfix}
\def\BibTeX{{\rm B\kern-.05em{\sc i\kern-.025em b}\kern-.08em
    T\kern-.1667em\lower.7ex\hbox{E}\kern-.125emX}}
\usepackage{eso-pic}

\definecolor{lightgray}{gray}{0.9}

\def\BibTeX{{\rm B\kern-.05em{\sc i\kern-.025em b}\kern-.08em
    T\kern-.1667em\lower.7ex\hbox{E}\kern-.125emX}}
\title{\LARGE \bf \vspace{6mm}
Robust Spatiotemporal Motion Planning for Multi-Agent Autonomous Racing via Topological Gap Identification and Accelerated MPC
}

\author{Mingyi Zhang$^{1*}$, Cheng Hu$^{1*}$, Yiqin Wang$^{1*}$, Haotong Qin$^{2\dagger}$, Hongye Su$^{1\dagger}$ and Lei Xie$^{1\dagger}$
\thanks{$^*$ \textbf{Equal contribution}.} 
\thanks{$1$ Authors are associated with the Department of Control Science and Engineering, Zhejiang University.}
\thanks{$2$ Authors are associated with the Center for Project-Based Learning, D-ITET, ETH Zurich.}
\thanks{$^\dagger$ Corresponding Authors {\tt\small {haotong.qin@pbl.ee.ethz.ch, lxie@iipc.zju.edu.cn}}.}}

\begin{document}

\newacronym{lidar}{LiDAR}{Light Detection and Ranging}
\newacronym{radar}{RADAR}{Radio Detection and Ranging}
\newacronym{mpc}{MPC}{Model Predictive Control}
\newacronym{mpcc}{MPCC}{Model Predictive Contouring Control}
\newacronym{iot}{IoT}{Internet of Things}
\newacronym{bev}{BEV}{Bird's-Eye View}

\newacronym{gpu}{GPU}{Graphics Processing Unit}
\newacronym{fps}{FPS}{Frames Per Second}
\newacronym{kf}{KF}{Kalman Filter}
\newacronym{ads}{ADS}{Autonomous Driving Systems}
\newacronym{iac}{IAC}{Indy Autonomous Challenge}
\newacronym{fsd}{FSD}{Formula Student Driverless}
\newacronym{rrt*}{RRT*}{Rapidly exploring Random Tree Star}
\newacronym{mgbt}{MGBT}{Multilayer Graph-Based Trajectory}
\newacronym{ftg}{FTG}{Follow The Gap}
\newacronym{gbo}{GBO}{Graph-Based Overtake}
\newacronym{rbf}{RBF}{Radial Basis Function}
\newacronym{map}{MAP}{Model- and Acceleration-based Pursuit}
\newacronym{sqp}{SQP}{Sequential Quadratic Programming}
\newacronym{roc}{RoC}{Region of Collision}
\newacronym{cpu}{CPU}{Central Processing Unit}
\newacronym{cots}{CotS}{Commercial off-the-Shelf}
\newacronym{obc}{OBC}{On-Board Computer}

\newacronym{sota}{SOTA}{State-of-the-Art}
\newacronym{qp}{QP}{Quadratic Programming}
\newacronym{A2RL}{A2RL}{Abu Dhabi Racing League}
\newacronym{GP}{GP}{Gaussian Process}
\newacronym{FITC}{FITC}{Fully Independent Training Conditional}
\newacronym{VFE}{VFE}{Variational Free Energy}
\newacronym{LQR}{LQR}{Linear Quadratic Regulator}
\newacronym{FSDP}{FSDP}{\textit{Fast and Safe Data-Driven Planner}}
\newacronym{gp}{GP}{Gaussian Process}
\newacronym{sgp}{SGP}{Sparse Gaussian Process}
\newacronym{VESC}{VESC}{Vedder Electronic Speed Controller}

\def\namealgo{RLPP}
\def\sim2real{Sim-to-Real}

\maketitle
\thispagestyle{fancy}

\begin{strip}
\vspace{-2.5cm}
\centering
\includegraphics[width=\textwidth]{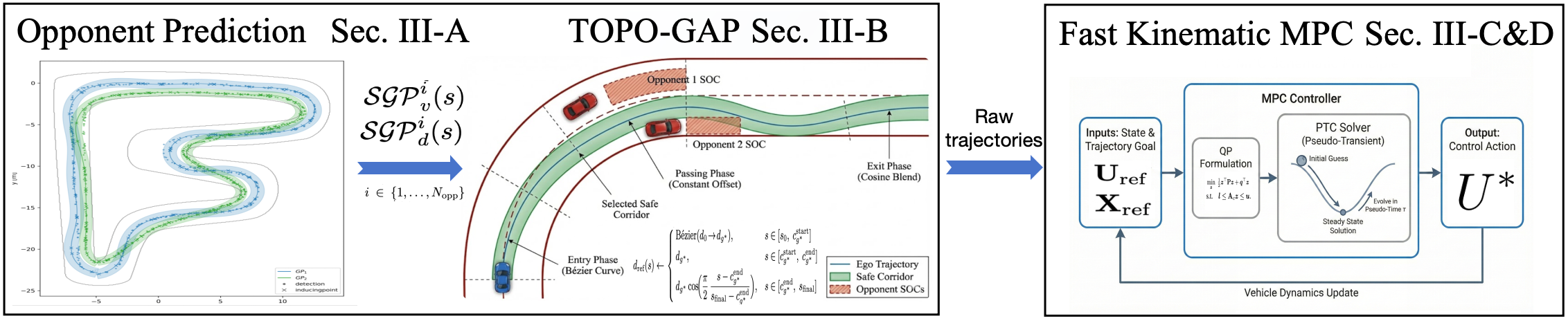}
\captionof{figure}{Framework of the proposed \emph{Topological Gap Identification Planner(Topo-Gap)} : The method first employs parallel \glspl{sgp} to predict multi-opponent spatiotemporal corridors. Based on these dynamic predictions, a topology-aware module identifies the optimal overtaking gap, while a PTC-accelerated \gls{mpc} optimizes the trajectory to ensure strict kinematic feasibility and safety at high speeds.}
\label{fig:graphical_abstract}
\vspace{-0.475cm}
\end{strip}

\pagestyle{empty}

\setlength{\tabcolsep}{3pt}
\glsresetall
\begin{abstract}
    High-speed multi-agent autonomous racing demands robust spatiotemporal planning and precise control under strict computational limits. Current methods often oversimplify interactions or abandon strict kinematic constraints. We resolve this by proposing a Topological Gap Identification and Accelerated MPC framework. By predicting opponent behaviors via SGPs, our method constructs dynamic occupancy corridors to robustly select optimal overtaking gaps. We ensure strict kinematic feasibility using a Linear Time-Varying MPC powered by a customized Pseudo-Transient Continuation (PTC) solver for high-frequency execution. Experimental results on the F1TENTH platform show that our method significantly outperforms state-of-the-art baselines: it reduces total maneuver time by 51.6\% in sequential scenarios, consistently maintains an overtaking success rate exceeding 81\% in dense bottlenecks, and lowers average computational latency by 20.3\%, pushing the boundaries of safe and high-speed autonomous racing.
\end{abstract}
\section{Introduction} \label{sec:intro}
Autonomous racing serves as a high-intensity proving ground for advancing autonomous driving technologies, pushing algorithms to their operational limits under extreme dynamic conditions.\cite{betz_weneed_ar, ar_survey}.

In head-to-head competitions, ranging from scaled platforms such as F1TENTH\cite{evans2024unifyingf1tenthautonomousracing} to full-scale events like the A2RL\cite{hoffmann2026headtoheadautonomousracinglimits} challenge, vehicles face the critical challenge of executing dynamic and complex overtaking maneuvers.
When navigating against multiple dynamic opponents, autonomous agents must rapidly exploit fleeting, spatiotemporally shifting gaps.
This demands generating optimal control sequences within milliseconds at the absolute limits of tire friction, forcing the planner to concurrently manage discrete topological decisions and strict kinematic feasibility under intense computational pressure.
In such edge-case scenarios, even marginal inaccuracies in kinematic planning or slight computational delays can precipitate catastrophic loss of control\cite{hoffmann2026headtoheadautonomousracinglimits}.

The methods for planning overtaking trajectories can be primarily divided into end-to-end direct trajectory generation and traditional hierarchical planning.
Although end-to-end methods based on reinforcement learning have shown promise in certain scenarios\cite{steiner2025acceleratingrealworldovertakingf1tenth}, the diverse strategies of real-world opponents make it difficult for these models to rapidly acquire interaction data and adapt online.
Consequently, hierarchical planning remains the prevailing option due to its rigorous safety guaranties\cite{Trauth_2024}.
Within this paradigm, single-opponent overtaking algorithms have matured significantly, successfully integrating data-driven predictions\cite{Kebbati_2024} (e.g., GP) to handle adversary uncertainty. However, while robust for one-on-one engagements, extending these predictive foundations to multi-vehicle head-to-head racing introduces entirely new dimensionalities of complexity.

1) \textbf{Dynamic and Complex Overtaking Windows:} Facing multiple opponents with diverse racing lines and speeds, overtaking gaps appear briefly and unpredictably. The planner must rapidly and accurately predict opponent behavior and seize these fleeting opportunities amid complex interactions;
2) \textbf{Kinematic Feasibility at the Limit:} Planned maneuvers must remain strictly within the bounds of vehicle dynamics; even minor violations at high speeds risk catastrophic instability, requiring robust kinematic constraints throughout the planning process.
3) \textbf{Strict Real-Time Constraints:} Millisecond-level response is essential, as computational delays directly erode viable overtaking opportunities in these high-speed, highly dynamic scenarios.

To tackle these challenges, this paper proposes a novel Topological Gap (Topo-Gap) overtaking framework with a fast QP-based MPC solver (Fig.~\ref{fig:graphical_abstract}). Given SGP-predicted opponent trajectories, our framework automatically constructs spatiotemporal overtaking corridors, identifies overtaking windows, generates adaptive trajectories, and leverages an accelerated MPC to optimize the control sequence for robust and accurate tracking.
The main contributions of this paper are summarized as follows:
 
\begin{enumerate}[I]
    \item 
    \textbf{SGPR-Based Dynamic Corridor Prediction:} A lightweight, probabilistic prediction framework based on parallel SGPs is proposed to generate dynamic, uncertainty-aware occupancy corridors for multiple opponents. These corridors enable robust identification of spatiotemporal overtaking windows and support adaptive trajectory planning.
    \item 
    \textbf{Topology-Aware, Hysteresis-Based Gap Selection:} A topology-driven hysteresis-augmented method is proposed to identify complex and rapidly changing overtaking gaps. Within this framework, the incorporation of a hysteresis cost function guarantees the stable selection of optimal safe gaps, thereby entirely suppressing decision oscillations during aggressive multi-agent competition.
    \item 
    \textbf{Robust PTC-Accelerated MPC:} PTC is employed to accelerate quadratic problem solving in MPC. To the best of our knowledge, this is the first time PTC has been applied to the domain of overtaking trajectory optimization. Combined with numerical stabilization via Tikhonov regularization and LDLT decomposition, our custom QP solver achieves consistently faster and more robust performance on dense QP problems than general-purpose solvers like OSQP, significantly reducing control latency and enhancing safety in highly dynamic conditions.
    \item
    \textbf{Open Source:} 
    The proposed planner is validated against state-of-the-art baselines in official F1TENTH simulation environment. The full codebase will be open-sourced upon the acceptance of this paper~\cite{forzaeth}.
\end{enumerate}

\section{Related Work} \label{sec:Rela_work}


\subsection{Multi-Agent Spatiotemporal Prediction}
Existing overtaking planners can be broadly classified into reactive and predictive approaches. 
Reactive methods, such as Follow-The-Gap (FTG) \cite{SEZER20121123}and sampling-based Frenet planners\cite{frenetplanner}, rely heavily on static assumptions, making them short-sighted in highly dynamic environments. 
To provide proactive collision awareness, recent State-of-the-Art (SOTA) methods integrate Gaussian Process Regression (GPR). 
M-PSpliner\cite{imholz2025mpredictivesplinerenablingspatiotemporal} successfully predicts multiple opponents using standard GPs; however, bounded by the $\mathcal{O}(N^3)$ complexity, it computes the Region of Collision (RoC) exclusively for the single closest opponent. 
This drastic simplification blinds the planner to the complex gaps formed by multiple interacting vehicles.
Conversely, FSDP\cite{hu2025fsdpfastsafedatadriven} mitigates the computational bottleneck via SGP, yet its framework is strictly confined to single-opponent scenarios.

Our planner addresses these limitations by deploying parallel SGPR models across all tracked opponents to maintain high-frequency global awareness. 
Based on this, we introduce a novel topological gap identification mechanism to proactively evaluate dynamic safety corridors and generate adaptive initial trajectories.
Consequently, this framework empowers the ego vehicle to seize fleeting multi-agent gaps and execute highly complex overtaking maneuvers, effectively resolving the myopic constraints of current SOTA methods.

\subsection{Optimization-Based Planning and Control} 
Recent predictive planners, ranging from Predictive Spliner\cite{baumann2024predictivesplinerdatadrivenovertaking} to its multi-agent extension M-PSpliner\cite{imholz2025mpredictivesplinerenablingspatiotemporal}, excel at computing proactive spatial evasion paths. 
However, because their trajectory generation relies purely on geometric optimization without considering vehicle kinematics, the resulting paths often exhibit poor executability at the absolute limits of handling.
Conversely, if a full vehicle dynamics model is directly embedded into the planning stage to ensure physical feasibility, it introduces prohibitive computational overhead and redundant calculations, severely violating the strict real-time constraints of high-speed racing.

Consequently, adopting a kinematic model within a MPC framework emerges as a highly effective compromise, bridging the gap between physical feasibility and real-time efficiency.
Although frameworks like FSDP\cite{hu2025fsdpfastsafedatadriven} successfully adopt this kinematic MPC paradigm, they typically rely on generic ADMM-based quadratic programming(QP) solvers like OSQP.
When subjected to the ill-conditioned matrices inherent to dense multi-agent constraints, these solvers suffer from slow tail convergence, which inevitably starves the high-frequency control loop. 

To resolve this bottleneck and robustly support the highly dynamic trajectories generated by our topology-aware framework, we introduce an engineered PTC\cite{10551407} solver into the racing LTV-MPC. 
By fortifying the standard PTC algorithm with practical enhancements like Tikhonov regularization and LDLT decomposition, our tailored solver efficiently handles near-singular matrices.
This adaptation significantly reduces computation time and ensures the deterministic, high-frequency execution required for aggressive multi-vehicle maneuvers.

\subsection{Summary of Existing Overtaking Algorithms}
\cref{tab:algorithm_comparison} summarizes the comparison between the method proposed in this paper and SOTA methods.

\begin{table}[!t]
    \centering
    \resizebox{\columnwidth}{!}{%
    \begin{tabular}{l|c|c|c|c}
    \toprule
    \textbf{Algorithm} & \textbf{Multi-Opp.} & \textbf{Kine. Feas.} & \textbf{Robust Solver} & \textbf{Topo. Gap} \\
    \midrule
    Frenet & Yes & No & No & No \\
    FTG & Yes & No & No & No \\
    M-PSpliner & \textbf{Yes} & No & No & No \\
    FSDP & No & \textbf{Yes} & No & No \\
    \textbf{Proposed (ours)} & \textbf{Yes} & \textbf{Yes} & \textbf{Yes} & \textbf{Yes} \\
    \bottomrule
    \end{tabular}%
    }
    \caption{Comparison of overtaking strategies. \textbf{Multi-Opp.}:Multiple opponent handling; \textbf{Kine. Feas.}:Strict kinematic restriction; \textbf{Robust Solver}:High-frequency accelerated solver; \textbf{Topo. Gap}:Dynamic gap identification. Ours uniquely satisfies all requirements.}
    \label{tab:algorithm_comparison}
    \end{table}
\section{Methodology} 
\label{sec:methodology}
In this section, we provide a detailed illustration of how the proposed multi-vehicle overtaking planner operates. 
First, we utilize parallel SGP models to predict the behaviors of multiple opponents and construct variance-inflated dynamic corridors.
Next, we demonstrate how to use our topology-aware gap selection mechanism to identify the optimal overtaking channel, and generate an initial kinematically adaptive trajectory through polynomial fitting and differential flatness techniques. 
Finally, an engineered PTC accelerated LTV-MPC enforces strict safety and kinematic feasibility at high speeds.
Figure~\ref{fig:graphical_abstract} shows the hierarchical structure of the planner.

\subsection{Spatiotemporal Prediction and Dynamic Corridors}
\label{ssec:corridor}

\subsubsection{Multi-Opponent Sparse GP Prediction} 
\begin{figure}[t]
    \centering
    
    \includegraphics[width=\columnwidth]{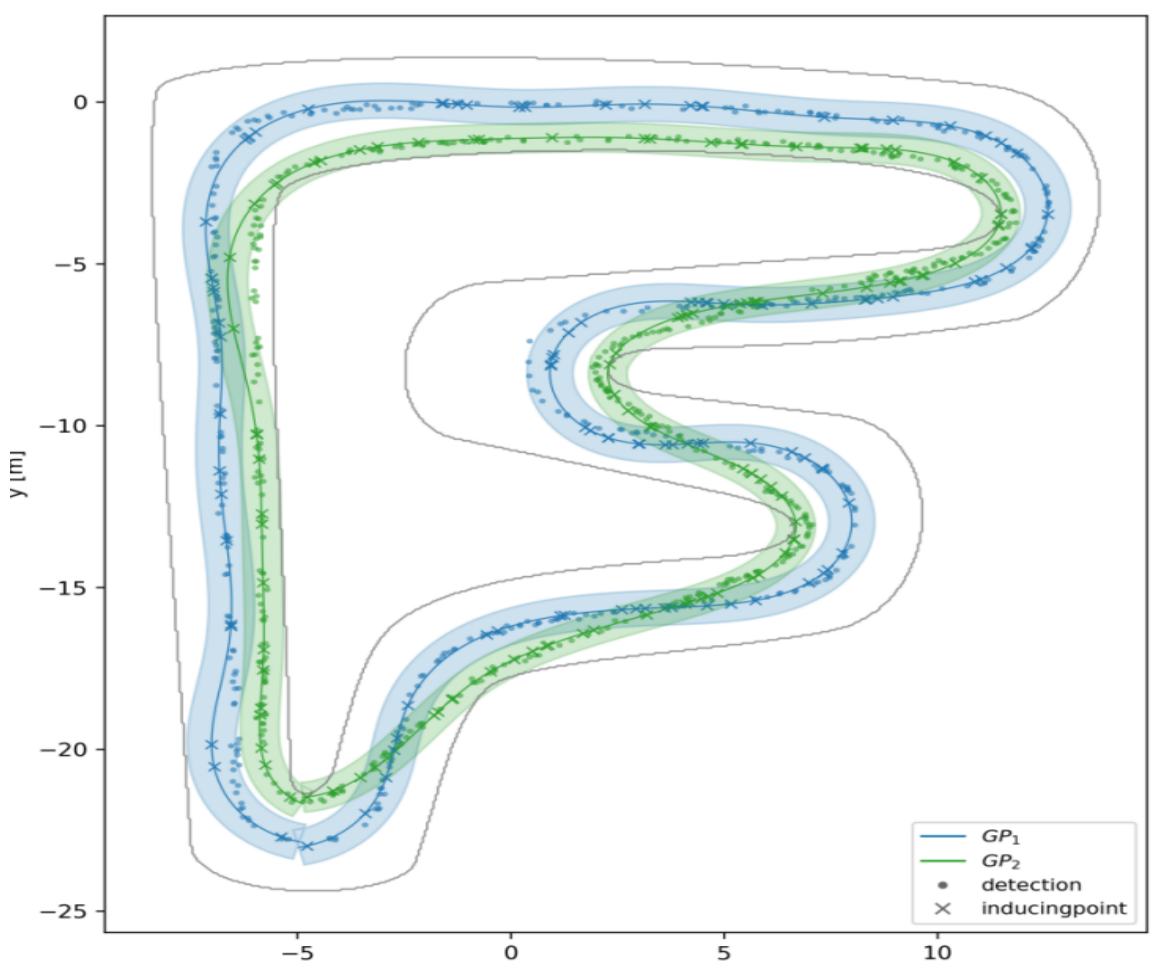}
    \caption{Qualitative visualization of the multi-opponent behavior prediction via parallel SGPs. The models accurately infer the continuous lateral deviation and velocity profiles, generating variance-inflated spatial bounds to support the downstream dynamic corridor construction.}
    \label{fig:sgp}
\end{figure}

To efficiently model opponent behaviors in the Frenet frame, we employ SGPs with $M$ inducing points ($M \ll N$). As depicted in Fig.~\ref{fig:sgp}, extracting these sparse inducing points from dense raw detections significantly reduces computational overhead while preserving dynamic spatial bounds. For opponent $i \in \{1,...,N_{opp}\}$, independent models predict the lateral deviation $d$ and velocity $v$ over arc-length $s$:
\begin{align}
    d^i(s) &\thicksim \mathcal{N}(\mu_d^i(s), \sigma_d^{i2}(s)), \\
    v^i(s) &\thicksim \mathcal{N}(\mu_v^i(s), \sigma_v^{i2}(s)).
\end{align}
These parallel $\mathcal{O}(NM^2)$ SGP instances ensure efficient and continuous uncertainty quantification for real-time inference.

\subsubsection{Spatiotemporal Forward Simulation for Multi-Opponent SOC}
Existing single-opponent methods typically output a single collision interval $[c_{\text{start}}, c_{\text{end}}]$ with a \emph{constant} lateral bound. 
Even in multi-opponent extensions, this often collapses to evaluating only the closest adversary. 
We instead compute a continuous, \emph{trajectory-following} representation for \emph{every} tracked opponent. 
This spatiotemporal corridor is parameterized by the ego vehicle's future arc-length $s$, explicitly mapping the opponent's predicted lateral occupancy to the exact instant the ego vehicle traverses $s$.

For each opponent $i \in \{1,\dots,N_{\text{opp}}\}$, we perform a time-domain forward integration over a horizon $T_{\text{hor}}$ with step $\Delta t$. The longitudinal states of the ego vehicle and the opponent evolve as:
\begin{equation}
\begin{aligned}
    s_{\text{ego}}(t{+}\Delta t) &= s_{\text{ego}}(t) + v_{\text{ego}}(t)\Delta t + \tfrac{1}{2}a(t)\Delta t^2, \\
    s^i_{\text{opp}}(t{+}\Delta t) &= s^i_{\text{opp}}(t) + \mu^i_v\!\bigl(s^i_{\text{opp}}(t)\bigr)\Delta t,
\end{aligned}
\label{eq:ego_opp_sim}
\end{equation}
where the ego acceleration $a(t)$ is scaled linearly between $a_{\min}$ and $a_{\max}$ based on the current velocity, and $\mu^i_v$ is the opponent's SGP-predicted mean velocity. 

The Spatiotemporal Occupancy Corridor (SOC) spans a longitudinal interaction window, denoted as $\mathcal{Z}^i = [c^i_{\text{start}}, c^i_{\text{end}}]$, which corresponds to the red hatched areas illustrated in Fig\ref{fig:soc_topo}. It is defined as the ego $s$-interval where the signed longitudinal distance $s^i_{\text{opp}} - s_{\text{ego}}$ strictly falls within $(-L_{\text{rear}}, L_{\text{front}})$. During this interaction window, the central data structure is the \emph{occupancy profile} $\mathcal{P}^i$, recorded at each time step $t_j$:
\begin{equation}
    \mathcal{P}^i = \Bigl\{\Bigl(s_{\text{ego}}(t_j),\; \mu^i_d\bigl(s^i_{\text{opp}}(t_j)\bigr),\; \sigma^{i\,2}_d\bigl(s^i_{\text{opp}}(t_j)\bigr)\Bigr)\Bigr\}_{j=1}^{N_{\text{soc}}}.
    \label{eq:occupancy_profile}
\end{equation}
Consequently, $\mathcal{P}^i$ forms an ego-$s$-parameterized sequence mapping the opponent's precise lateral mean and variance to the ego vehicle's arrival coordinate. Instead of aggregating into two endpoints, this full profile drives the subsequent corridor construction, avoiding the over-conservative abstraction of static bounding boxes.

\subsubsection{Variance-Inflated Corridor Construction}
Leveraging $\mathcal{P}^i$, we define the dynamic corridor $\mathcal{C}^i$ with spatially varying lateral boundaries. At each sampled coordinate $s_k$ in the profile, the effective width intelligently incorporates the GP predictive variance to ensure a probabilistic safety margin:
\begin{equation}
\begin{split}
    W_{\text{eff}}(s_k) = \min \Bigl( &W_{\text{car}} + W_{\text{margin}} \\
    &+ k_\sigma\sqrt{\sigma^{i\,2}_d(s_k)},\; W_{\max} \Bigr),
\end{split}
\label{eq:W_eff}
\end{equation}
where $W_{\text{car}}$ is the vehicle width, $W_{\text{margin}}$ is the base evasion clearance, $k_\sigma$ controls the confidence interval (e.g., $k_\sigma{=}2$ for ${\approx}95\%$ confidence), and $W_{\max}$ caps the extreme width. The nominal lateral boundaries are thus defined as:
\begin{equation}
\begin{aligned}
    d^i_L(s_k) &= \mu^i_d(s_k) + \tfrac{1}{2} W_{\text{eff}}(s_k), \\
    d^i_R(s_k) &= \mu^i_d(s_k) - \tfrac{1}{2} W_{\text{eff}}(s_k).
\end{aligned}
\label{eq:corridor_bounds}
\end{equation}
To ensure the reliability of the subsequent topological gap identification, we must eliminate artificial geometric pinch points (i.e., transient narrowings) caused by high-frequency variance jitter in the SGP predictions.
We apply a 1D spatial morphological dilation over a longitudinal window $\Delta s_w$, enforcing a monotonically safe and smooth envelope:
\begin{equation}
\begin{aligned}
    \tilde{d}^i_L(s_k) &= \max_{s_j \in \left[ s_k - \frac{\Delta s_w}{2}, \, s_k + \frac{\Delta s_w}{2} \right]} d^i_L(s_j), \\
    \tilde{d}^i_R(s_k) &= \min_{s_j \in \left[ s_k - \frac{\Delta s_w}{2}, \, s_k + \frac{\Delta s_w}{2} \right]} d^i_R(s_j).
\end{aligned}
\label{eq:convex_envelope}
\end{equation}

Furthermore, when an opponent executes a rapid lane change, the standard variance inflation combined with the lateral displacement projects an overly conservative diagonal barrier across the track. To mitigate this, we extract the opponent's average lateral velocity over the interaction window $\Delta t_{\text{soc}} = t_{N_{\text{soc}}} - t_1$:
\begin{equation}
    \bar{v}^i_d = \frac{\mu^i_d\bigl(s^i_{\text{opp}}(t_{N_{\text{soc}}})\bigr) - \mu^i_d\bigl(s^i_{\text{opp}}(t_1)\bigr)}{\Delta t_{\text{soc}}}.
    \label{eq:crossing_vel}
\end{equation}
If the maneuver is classified as crossing ($|\bar{v}^i_d| > v_{d,\text{thresh}}$), the probabilistic inflation in \eqref{eq:W_eff} is bypassed. Instead, a deterministic tight bound $W^{\text{cross}}_{\text{eff}} = W_{\text{car}} + \eta W_{\text{margin}}$ (with $\eta \in (0,1)$) is enforced. This adaptive mechanism prevents transient lane-changes from projecting artificial wal that invalidate laterally adjacent, physically feasible overtaking gaps.

\subsection{Topology-Aware Gap Identification and Selection}
\label{ssec:topo_gap}
The construction of the dynamic corridors $\mathbf{C} = \{\mathcal{C}^1, \dots, \mathcal{C}^{N_{\text{opp}}}\}$ inherently partitions the drivable track into a finite set of topologically distinct free-space channels. 
Unlike reactive single-opponent planners that rely on simple heuristics to choose a binary overtaking side (e.g.,left or right), our framework systematically evaluates the complete free-space topology defined by $\mathbf{C}$ to identify the optimal passing channel. This is achieved through a systematic three-stage process: (1) spatial enumeration of all topologically valid gaps along the occupancy profile; (2) cost-based optimal channel selection fortified by decision hysteresis; and (3) generation of a kinematically feasible initial trajectory localized within the selected gap.
\subsubsection{Multi-Point Gap Sampling and Classification}
\label{sssec:gap_class}
To ensure continuous passability amidst opponents' lateral motions, we uniformly sample $N_s$ longitudinal coordinates $s_k \in [s_{\text{soc}}^{\min}, s_{\text{soc}}^{\max}]$ across the union of all interaction windows. At each $s_k$, the dynamic corridor boundaries are queried, and obstacles are laterally sorted in ascending order to form an ordered set $\mathcal{O}(s_k) = \{O_1, \dots, O_{N_{\text{opp}}}\}$. 

The available free space is systematically classified into Left, Right, and Middle gaps , with their respective widths computed as:
\begin{equation}
\begin{aligned}
    W_L(s_k) &= d^{\text{track}}_L(s_k) - \tilde{d}^{N_{\text{opp}}}_L(s_k) - \epsilon_{\text{side}}, \\
    W_R(s_k) &= \tilde{d}^{1}_R(s_k) - d^{\text{track}}_R(s_k) - \epsilon_{\text{side}}, \\
    W_{M,j}(s_k) &= \tilde{d}^{j+1}_R(s_k) - \tilde{d}^{j}_L(s_k) - 2\epsilon_{\text{mid}}.
\end{aligned}
\label{eq:gap_widths}
\end{equation}

A candidate gap $g$ is declared \emph{passable} if and only if its minimum width across all sampled points strictly exceeds a safe clearance threshold $W_{\min}$:
\begin{equation}
    \min_{k \in \{1,\dots,N_s\}} W_g(s_k) > W_{\min}.
    \label{eq:passable}
\end{equation}
This strict multi-point criterion inherently filters out channels that are transiently wide but dynamically close as the ego vehicle approaches.
\subsubsection{Cost-Based Gap Selection with Hysteresis}
\label{sssec:gap_select}
To suppress high-frequency decision oscillation inherent to greedy multi-agent planners, passable gaps $g \in \mathcal{G}$ are evaluated using a hysteresis-augmented cost function:
\begin{multline}
    J(g) = \frac{w_s}{\min_k W_g(s_k)} + w_r \,\bigl|d_{\text{center}}(g) - d_{\text{raceline}}\bigr| \\
    + w_c\, C_{\text{switch}}(g).
\label{eq:gap_cost}
\end{multline}
This formulation optimally balances the bottleneck clearance of the gap ($\min_k W_g(s_k)$) and the raceline tracking error against a discrete switching penalty $C_{\text{switch}}$:
\begin{equation}
    C_{\text{switch}}(g) = 
    \begin{cases}
        0, & \text{if } g = g_{\text{prev}}, \\
        C_0, & \text{if same side as } g_{\text{prev}}, \\
        C_0 + C_1, & \text{if opposite side}.
    \end{cases}
    \label{eq:switch_cost}
\end{equation}
Here, the heavy penalty $C_1$ explicitly discourages hazardous cross-corridor maneuvers at high speeds. 

To ensure strict temporal consistency, the active target gap $g^*$ is updated to a new candidate $g_{\text{new}}$ only if it yields a decisive cost reduction:
\begin{equation}
    J(g_{\text{new}}) < (1 - \alpha)\, J(g^*),
    \label{eq:hysteresis}
\end{equation}
where $\alpha \in (0,1)$ is the hysteresis margin. 
Moreover, this condition is gated by a discrete dwell timer that enforces a minimum holding period between consecutive switches, strictly guaranteeing the kinematic stability of the downstream framework.
\begin{figure}[htbp]
    \centering
    \includegraphics[width=\columnwidth]{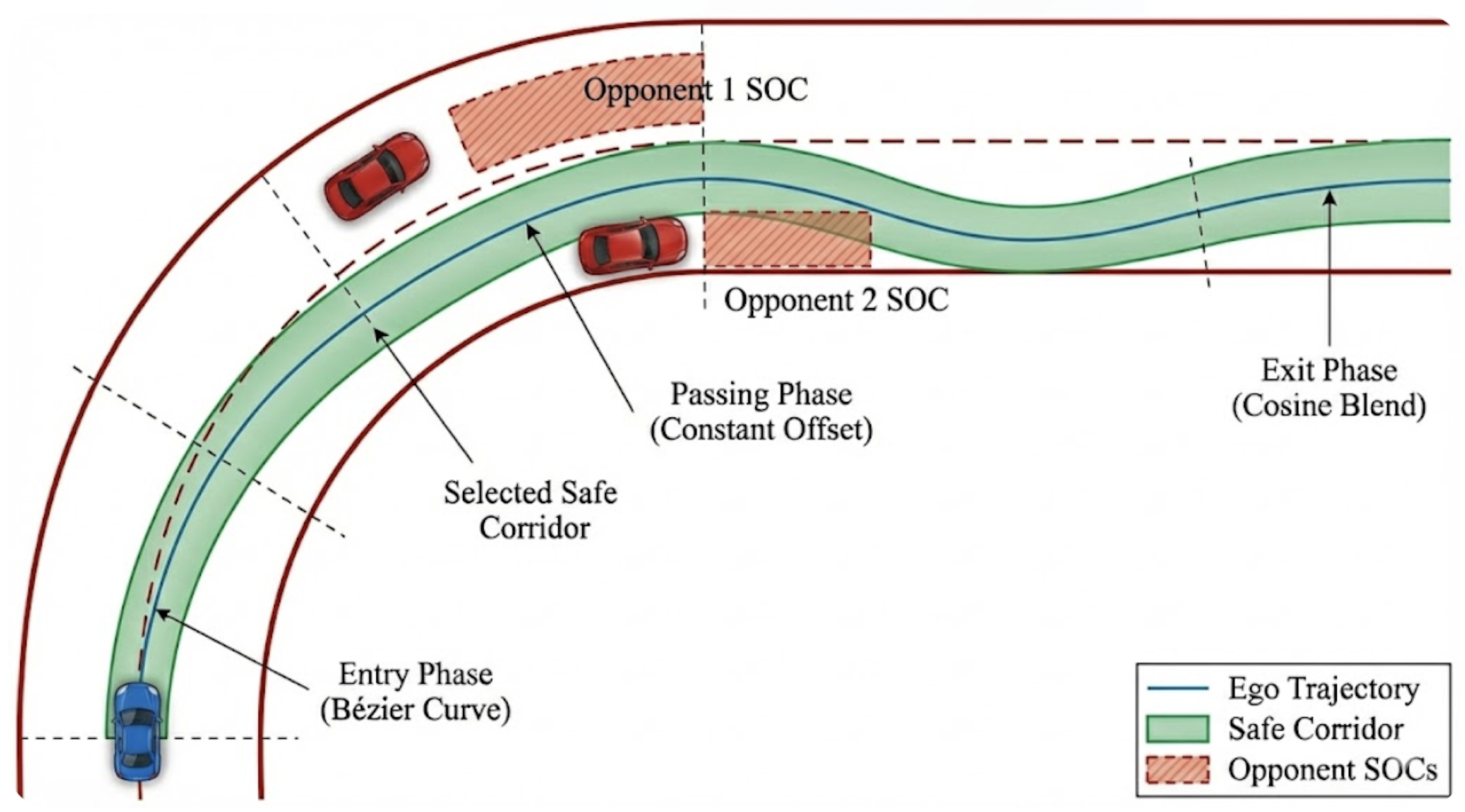}
    \caption{Illustration of the topology-aware gap selection and adaptive trajectory generation. The ego vehicle (blue) navigates a continuous safe corridor (green) to cleanly bypass the opponents' Spatiotemporal Occupancy Corridors (SOCs, red hatched areas) utilizing a three-phase geometric path.}
    \label{fig:soc_topo}
\end{figure}

\subsubsection{Adaptive Initial Trajectory Generation}
\label{sssec:init_traj}
Once the optimal target gap $g^*$ and its lateral offset $d_{\text{target}}$ are identified (highlighted in green in Fig.~\ref{fig:soc_topo}), we synthesize a piecewise-smooth geometric path $d_{\text{ref}}(s)$ strictly confined within the safe corridor. This path provides a kinematically favorable initialization for the downstream solver through three distinct phases:

\begin{itemize}
    \item \textbf{Entry Phase:} A cubic Bézier curve steers the vehicle from $d_0$ to $d_{\text{target}}$. To prevent hazardous late cut-ins, the control points ($\gamma_1 {\approx} 0.35, \gamma_2 {\approx} 0.88$) are heavily biased to enforce an early lateral commitment before reaching the opponents' SOCs.
    
    \item \textbf{Passing Phase:} The trajectory maintains a constant offset $d_{\text{target}}$, cleanly traversing the longitudinal span of the interaction window.
    
    \item \textbf{Exit Phase:} A $C^2$-continuous cosine blend, defined as $d(s) = d_{\text{target}} \cos\!\big(\frac{\pi}{2} \frac{s - s_{\text{end}}}{s_{\text{final}} - s_{\text{end}}}\big)$, smoothly merges the vehicle back to the optimal raceline once clear of the final SOC.
\end{itemize}

Finally, $d_{\text{ref}}(s)$ is clamped within the physical track boundaries. The complete topology-aware planning pipeline is summarized in Algorithm~\ref{algo:topo_gap}.
\begin{algorithm}[htbp]
    \small
    \caption{Topology-Aware Gap Selection \& Reference Generation}
    \label{algo:topo_gap}
    \begin{algorithmic}[1]
    \State \textbf{Input:} $\{\mathcal{SGP}^i\}$, $\partial\mathcal{D}$, $(s_0,d_0)$, $g^*_\text{prev}$, $T_\text{dwell}$ \enspace \textbf{Out:} $d_\text{ref}(s)$, $g^*$
    \Statex \textbf{Corridor Construction:}
    \For{$i = 1 \to N_\text{opp}$}
        \State Integrate \eqref{eq:ego_opp_sim}$\;\Rightarrow\;\mathcal{P}^i$, $\;\mathcal{Z}^i{=}[c^i_s,c^i_e]$
        \State $(k_\sigma,W_\text{m}) \!\gets\! \begin{cases}(0,\;\eta W_\text{mg}) & |\bar{v}^i_d|>v_\text{thr}\\(k_\sigma,\;W_\text{mg}) & \text{else}\end{cases}$
        \State $W_\text{eff}(s_k) \gets \min(W_\text{car}{+}W_\text{m}{+}k_\sigma\sigma^i_d,\,W_\text{max})$\Comment{\eqref{eq:W_eff}}
        \State $\tilde{d}^i_{L/R} \gets \operatorname{Dilate}(\mu^i_d{\pm}\tfrac{1}{2}W_\text{eff},\,\Delta s_w)$\Comment{\eqref{eq:corridor_bounds},\eqref{eq:convex_envelope}}
    \EndFor
    \Statex \textbf{Gap Identification:}
    \State Sample $\{s_k\}^{N_s}$; compute $\{W_L,W_R,W_{M,j}\}$ via \eqref{eq:gap_widths}
    \State $\mathcal{G} \gets \{g \mid \min_k W_g(s_k) > W_\text{min}\}$
    \Statex \textbf{Gap Selection:}
    \State $g_\text{new} \gets \arg\min_{g\in\mathcal{G}}\,\bigl[\tfrac{w_s}{\min_k W_g} + w_r|d_g {-} d_\text{rl}| + w_c C_\text{sw}\bigr]$\Comment{\eqref{eq:gap_cost}}
    \If{$J(g_\text{new}) {<} (1{-}\alpha)J(g^*)$ \textbf{and} $T_\text{el}{>}T_\text{dwell}$}\Comment{\eqref{eq:hysteresis}}
        \State $g^* \gets g_\text{new}$ 
    \Else\enspace $g^* \gets g^*_\text{prev}$ 
    \EndIf
    \Statex \textbf{Reference Synthesis:}
    \State $d_\text{ref}(s) \gets$
$\begin{cases}
\text{B\'{e}zier}(d_0 \!\to\! d_{g^*}), & s \in [s_0,\;c^\text{start}_{g^*}] \\[2pt]
d_{g^*}, & s \in [c^\text{start}_{g^*},\;c^\text{end}_{g^*}] \\[2pt]
d_{g^*}\cos\!\Bigl(\dfrac{\pi}{2}\,\dfrac{s - c^\text{end}_{g^*}}{s_\text{final} - c^\text{end}_{g^*}}\Bigr), & s \in [c^\text{end}_{g^*},\;s_\text{final}]
\end{cases}$
\State $d_\text{ref} \gets \operatorname{clip}\!\bigl(d_\text{ref},\;\partial\mathcal{D}\bigr)$\Comment{enforce track boundaries}
\end{algorithmic}
\parbox{\columnwidth}{\footnotesize \noindent \textit{Note:} $\partial\mathcal{D}$ denotes the boundaries of the race track.}
\end{algorithm}

\subsection{Polynomial Fitting and Reference State Extraction}
\label{ssec:diff_flat}
To provide a kinematically feasible initial state for the downstream LTV-\gls{mpc}, the adaptive geometric path $d_{\text{ref}}(s)$ generated in the previous stage is fitted to a time-parameterized quintic polynomial. Following the bi-level paradigm established in FSDP~\cite{hu2025fsdpfastsafedatadriven}, this polynomial is converted to Cartesian coordinates. By exploiting the differential flatness property of the kinematic bicycle model with flat outputs $\mathbf{z} = [x, y]^\top$, the full reference state trajectory $\mathbf{X}_{\text{ref}}$ and control inputs $\mathbf{U}_{\text{ref}}$ are analytically recovered. These directly serve as the exact linearization points for the subsequent optimization.
\subsection{Robust LTV-MPC via PTC-Accelerated QP Solver}
\label{ssec:mpc}
To guarantee strict kinematic feasibility and safety at extreme speeds, the initial reference trajectory is refined via a MPC formulated in the Frenet frame. The vehicle dynamics are described by the kinematic bicycle model with state vector $\bm{x} = [s, d, \Delta\theta]^\top$ and input vector $\bm{u} = [v, \delta]^\top$, where $s$ is the arc length, $d$ is the lateral deviation, and $\Delta\theta$ is the heading error.

To formulate a computationally tractable optimization problem, the highly nonlinear vehicle dynamics are discretized and linearized around the reference states $\bm{X}_{\text{ref}}$ and inputs $\bm{U}_{\text{ref}}$ extracted from the preceding differential flatness module. This yields a Linear Time-Varying (LTV) system $\bm{x}_{k+1} = \mathbf{A}_k \bm{x}_k + \mathbf{B}_k \bm{u}_k$.

The MPC objective minimizes the quadratic tracking deviation $\sum (\|\bm{x}_k - \bm{x}_{k,\text{ref}}\|_{\mathbf{Q}}^2 + \|\bm{u}_k - \bm{u}_{k,\text{ref}}\|_{\mathbf{R}}^2)$, subject to the vehicle's actuation limits. Crucially, the multi-opponent spatiotemporal corridors generated in the topology-aware layer are interpolated as strict spatial bounds on the predicted lateral deviation, enforcing $d_{\min}(s_k) \le d_k \le d_{\max}(s_k)$ at each prediction step.

By condensing the state sequence through the linear dynamics, we analytically eliminate the state variables and reduce the optimization to a dense Quadratic Program (QP) strictly over the control input sequence $\bm{z}$:
\begin{equation}
    \min_{\bm{z}} \;\tfrac{1}{2}\bm{z}^\top\mathbf{P}\bm{z} + \bm{q}^\top\bm{z}
    \quad\text{s.t.}\quad \bm{l} \le \mathbf{A}_c\bm{z} \le \bm{u}.
    \label{eq:dense_qp}
\end{equation}
At racing speeds, near-singular $\mathbf{P}$ caused by tightly packed corridors renders standard solvers (e.g., OSQP) prohibitively slow.
We are the first to deploy PTC~\cite{10551407} in a multi-opponent overtaking context, exploiting its KKT-to-dynamics reformulation for guaranteed real-time performance.

\noindent\textbf{PTC Core.}
Partitioning $\mathbf{A}_c$ into equalities $(\mathbf{C},\bm{p})$ and inequalities $(\mathbf{D},\bm{q}_\text{ineq})$, the KKT optimality conditions reduce analytically to a single equation in the dual inequality multiplier $\bm{\lambda}\!\ge\!\bm{0}$:
\begin{equation}
    F(\bm{\lambda}) \triangleq \mathbf{G}\bm{\lambda} + \bm{h} - \varphi\!\bigl(\mathbf{G}\bm{\lambda}+\bm{h}\bigr) = \bm{0},
    \label{eq:ptc_F}
\end{equation}
where $\varphi(\bm{x})\!=\!\min(\bm{x},\bm{q}_\text{ineq})$ (component-wise), and the reduced matrices are:
\begin{equation}
\begin{aligned}
    \mathbf{G}' &= \mathbf{P}^{-1} - \mathbf{P}^{-1}\mathbf{C}^\top\!(\mathbf{C}\mathbf{P}^{-1}\mathbf{C}^\top)^{-1}\mathbf{C}\mathbf{P}^{-1},\\
    \bm{h}' &= \mathbf{P}^{-1}\!\bigl(\mathbf{C}^\top\!(\mathbf{C}\mathbf{P}^{-1}\mathbf{C}^\top)^{-1}\!(\mathbf{C}\mathbf{P}^{-1}\bm{q}+\bm{p})-\bm{q}\bigr),\\
    \mathbf{G} &= \mathbf{D}\mathbf{G}'\mathbf{D}^\top, \quad \bm{h} = \mathbf{D}\bm{h}'.
\end{aligned}
\label{eq:ptc_reduced}
\end{equation}
PTC solves \eqref{eq:ptc_F} by integrating $\dot{\bm{\lambda}} = -\beta F(\bm{\lambda})$, whose equilibrium $\bm{\lambda}^*$ is proven globally asymptotically stable (GAS) via Lyapunov theory~\cite{10551407}.
The primal solution is then recovered in closed form as $\bm{z}^* = \mathbf{G}'\mathbf{D}^\top\bm{\lambda}^* + \bm{h}'$.

\noindent\textbf{Engineering Adaptations.}
\textbf{1) Dual-Level Tikhonov Regularization.}
Racing corridors cause near-singular $\mathbf{P}$ and ill-conditioned Schur complements, destabilizing the \texttt{LDLT} factorizations in \eqref{eq:ptc_reduced}. We inject regularization at both levels: $\mathbf{P}_\text{reg}=\mathbf{P}+\epsilon_1\mathbf{I}$ and $\mathbf{C}\mathbf{P}^{-1}_\text{reg}\mathbf{C}^\top+\epsilon_2\mathbf{I}$, preserving solution quality while ensuring numerical stability.

\noindent\textbf{2) Fixed-Budget Integration \& Safe Fallback.}
The dynamics are discretized via explicit Euler:
\begin{equation}
    \bm{\lambda}^{(k+1)} = \bm{\lambda}^{(k)} - h\beta\,F\!\bigl(\bm{\lambda}^{(k)}\bigr),
    \label{eq:ptc_update}
\end{equation}
hard-capped at $K_{\max}\!=\!200$ iterations to guarantee $\mathcal{O}(1)$ timing. If extreme constraint tightness drives $\bm{\lambda}$ to \texttt{NaN}, the solver immediately resets $\bm{\lambda}\!\gets\!\bm{0}$ and exits, ensuring the chassis always executes an unconstrained-optimal command rather than failing.
\section{Experimental Results}
\label{sec:exp_results}

This section evaluates the proposed Topo-Gap method against state-of-the-art baselines across three dimensions: multi-opponent overtaking robustness, trajectory kinematic quality, and solver efficiency.

\subsection{Experimental Setup}
\label{subsec:exp_setup}
For a rigorous and fair comparison, all evaluations are conducted within a unified 1:10 scaled F1TENTH simulation environment, where every planner utilizes identical track maps and a standard Pure Pursuit\cite{Becker_2023}trajectory tracking controller.Computational benchmarking is performed on an Intel Core Ultra 9 275HX processor (3.1\,GHz).

\subsubsection{Multi-Opponent Scenario Definitions}
To systematically stress-test multi-agent planning capabilities, we designed four distinct traffic layouts of increasing topological complexity:
\begin{enumerate}[a)]
    \item \textbf{Sequential Mixed-Path (SMP):} Opponents follow the centerline and shortest path with a $20\,\mathrm{m}$ longitudinal gap, establishing a baseline for consecutive, independent overtakes.
    \item \textbf{Dense Same-Lane (DSL):} Two opponents travel on the optimal racing line with a tight $3\,\mathrm{m}$ headway, testing the planner's capacity to bypass continuous platoon blockages.
    \item \textbf{Tight Staggered Overlap (TSO):} Opponents drive almost side-by-side ($1\,\mathrm{m}$ gap) across the racing line and centerline, forming a severe parallel bottleneck.
    \item \textbf{Offset Needle Thread (ONT):} Opponents are diagonally staggered , evaluating the algorithm's robustness in threading dynamic narrow gaps without decision oscillation.
\end{enumerate}
\begin{figure}[t]
    \centering
    \includegraphics[width=\columnwidth]{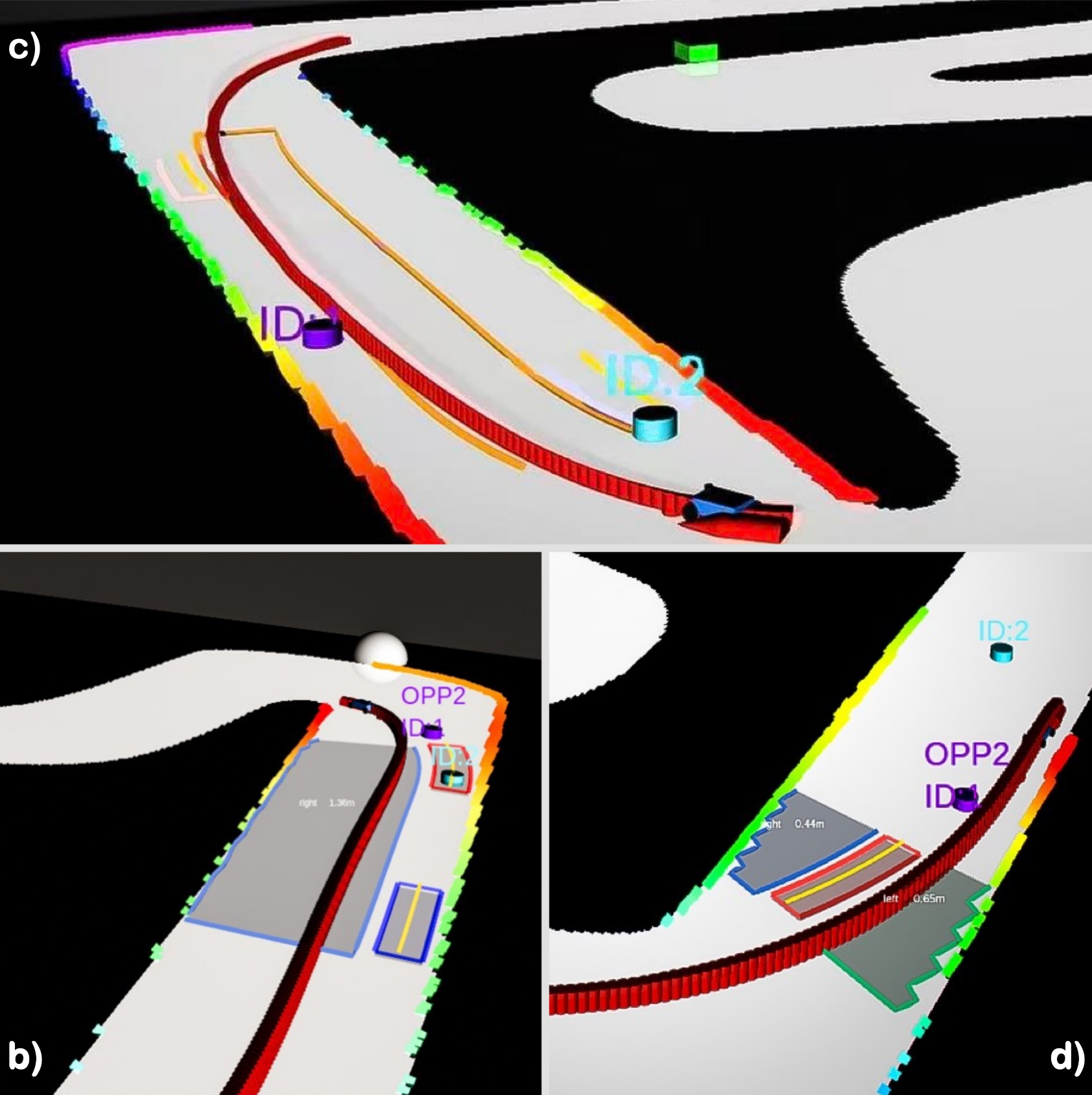}
    \caption{Overtaking maneuvers demonstrated in the F1TENTH RViz simulation environment across scenarios (b), (c), and (d). By utilizing the proposed TopoGap planner, the ego vehicle efficiently identifies navigable gaps (indicated by long rectangles) among the opponents' SOCs(denoted by short rectangles) and generates an optimal, smooth trajectory to safely execute the overtakes.}
    \label{fig:overtake}
\end{figure}
\begin{figure*}[!t]
    \centering
    \includegraphics[width=\textwidth]{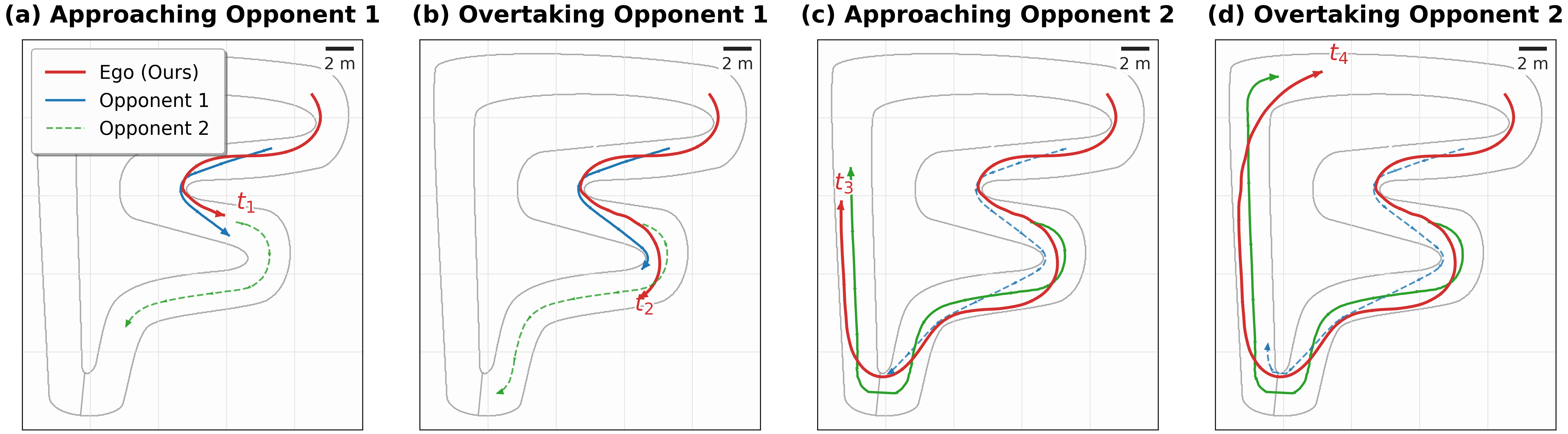}
    \caption{Qualitative overview of the proposed Topo-Gap framework executing a multi-opponent overtaking maneuver under the \textbf{SMP} scenario. (a) At $t_1$, the ego vehicle (magenta) trails Opponent 1 (gray) and evaluates the spatial gap for an outside overtake. (b) At $t_2$, the ego vehicle successfully clears the first opponent at the curve apex. (c) At $t_3$, it approaches Opponent 2 (teal) while continuously projecting spatiotemporal occupancy corridors. (d) At $t_4$, a safe and smooth inside overtake is completed. The generated trajectories strictly maintain kinematic feasibility without decision oscillation during the highly dynamic multi-agent interactions.}
    \label{fig:qualitative_overtake}
\end{figure*}
\subsubsection{Evaluation Metrics}
Each trial is evaluated until 5 successful overtakes are achieved. The metrics target distinct aspects of overtaking quality:
\begin{itemize}
    \item $\mathcal{R}_{\mathrm{ot/c}} [\%]$: Overtaking success rate (collision-free completions vs. total attempts).
    \item $\mathcal{S}_{\max} [\%]$: Maximum feasible speed scaler, defined as the unobstructed lap time ratio $\mathcal{T}_{ego}/\mathcal{T}_{opp}$ (where $\mathcal{T}_{ego}$ and $\mathcal{T}_{opp}$ are the baseline lap times of the ego and opponent, respectively). This metric is evaluated by progressively increasing the opponent's speed to determine the upper limit at which the ego can still consistently execute stable overtakes without losing control.
    \item $\Sigma\mathcal{T} [\mathrm{s}]$: Total maneuver time. This is strictly defined as the continuous duration from the moment the ego vehicle initiates tracking of the first opponent until it completely clears the final opponent and safely merges back onto the global optimal racing line.
    \item $\bar{\dot{a}},\, \bar{\dot{\delta}}$: Mean absolute jerk $[\mathrm{m/s^3}]$ and steering rate $[\mathrm{rad/s}]$, quantifying kinematic smoothness.
\end{itemize}

\subsubsection{GP Computational Overhead Analysis }
We compare the prediction efficiency of Topo-Gap and M-PSpliner by enforcing a unified execution framework. To ensure an unbiased evaluation, both planners share the exact same trigger rules, binning parameters, waypoint configurations, and prediction policies. Workloads are handled by dispatching one asynchronous task per tracked opponent into a multiprocessing pool of size 2.The reported benchmarks specifically evaluate the mean computational times dedicated to both GP training and inference across all valid updates.

\subsubsection{Solver Benchmarking Protocol}
To rigorously isolate and compare pure optimization efficiency, we employ a \textit{shadow mode} execution. At each control step, both our PTC-accelerated solver and the baseline OSQP solver receive and process the exact same QP problem formulated by the MPC. However, only the PTC commands are published to the vehicle chassis. This guarantees perfectly synchronized dynamic states, enabling a direct, head-to-head comparison of computation latency without compounding state divergence.

\subsection{Simulation Results}
\label{subsec:sim_results}

\subsubsection{Multi-Opponent Overtaking Performance}
As detailed in Table \ref{tab:multi_scenario_results}, both planners are evaluated across the four multi-agent topologies. In the SMP layout (Scenario A), both methods successfully complete the overtakes. While M-PSpliner achieves a comparable maximum speed scaler ($\mathcal{S}_{\max}$ of $70.2\%$ vs. $68.6\%$), Topo-Gap completes the maneuver more efficiently, reducing the total maneuver time $\Sigma\mathcal{T} $ from $35.75\,\mathrm{s}$ to $17.32\,\mathrm{s}$. A qualitative overview of our planner executing this highly dynamic multi-agent interaction is visually demonstrated in Fig. \ref{fig:qualitative_overtake}.

In more tightly coupled topologies, differences in obstacle handling strategies become apparent. The baseline M-PSpliner, which formulates the Region of Collision based on the single closest opponent, encounters challenges in the DSL platoon (Scenario B), yielding a success rate ($\mathcal{R}_{\mathrm{ot/c}}$) of only $45.5\%$. In contrast, Topo-Gap evaluates the extended multi-opponent corridor(as demonstrated in Fig. \ref{fig:overtake}b), achieving a $93.8\%$ success rate in $16.33\,\mathrm{s}$. 

In the TSO bottleneck (Scenario C), the baseline exhibits decision oscillation, leading to an extended total maneuver time of $88.0\,\mathrm{s}$. As visually demonstrated by the simulation results in Fig. \ref{fig:overtake}c, Topo-Gap leverages simultaneous Spatiotemporal Occupancy Corridors to decisively resolve the topological gap, safely executing the maneuver in just $8.30\,\mathrm{s}$. Furthermore, in the ONT layout (Scenario D), M-PSpliner struggles to find a feasible path, whereas Topo-Gap successfully navigates the dynamic narrow corridor(illustrated in Fig. \ref{fig:overtake}d in $24.20\,\mathrm{s}$ with an $83.3\%$ success rate.

\begin{table}[htbp]
    \centering
    \caption{Quantitative Overtaking Performance Across Multi-Opponent Scenarios}
    \label{tab:multi_scenario_results}
    \resizebox{\columnwidth}{!}{%
    \begin{tabular}{l | l c c c c}
    \toprule
    \textbf{Scen.} & \textbf{Planner} & \bm{$\mathcal{R}_{\mathrm{ot/c}} [\%]$} & \bm{$\Sigma\mathcal{T} [\mathrm{s}]$} & \bm{$\bar{\dot{a}} [\mathrm{m/s^3}]$} & \bm{$\bar{\dot{\delta}} [\mathrm{rad/s}]$} \\
    \midrule
    \multirow{2}{*}{\textbf{SMP}} 
    & M-PSpliner & 71.4  & 35.75 & 39.60 & 0.76 \\
    & \textbf{Ours} & \textbf{81.3}& \textbf{17.32} & \textbf{22.70} & \textbf{0.72} \\
    \midrule
    \multirow{2}{*}{\textbf{DSL}} 
    & M-PSpliner & 45.5 & 24.12 & 62.80 & 1.05 \\
    & \textbf{Ours} & \textbf{93.8}& \textbf{16.33} & \textbf{32.20} & \textbf{0.62} \\
    \midrule
    \multirow{2}{*}{\textbf{TSO}} 
    & M-PSpliner & 42.9 & 88.00 & 57.10 & 0.64 \\
    & \textbf{Ours} & \textbf{81.8}  & \textbf{8.30} & \textbf{40.74} & \textbf{0.55} \\
    \midrule
    \multirow{2}{*}{\textbf{ONT}} 
    & M-PSpliner & -- & Fail & 72.70 & 0.91 \\
    & \textbf{Ours} & \textbf{83.3} & \textbf{24.20} & \textbf{51.80} & \textbf{0.76} \\
    \bottomrule
    \end{tabular}%
    }
    \vspace{-1.5em}
\end{table}

\subsubsection{Kinematic Feasibility and Trajectory Quality}
The kinematic quality of the generated trajectories is evaluated using mean absolute jerk ($\bar{\dot{a}}$) and steering rate ($\bar{\dot{\delta}}$). The baseline's sequential geometric splicing results in severe variations in control commands, peaking at an average jerk of $72.7\,\mathrm{m/s^3}$ (Scenario D) and a steering rate of $1.05\,\mathrm{rad/s}$ (Scenario B). These rapid changes pose significant tracking challenges for low-level controllers at extreme speeds. By integrating a kinematic MPC, Topo-Gap explicitly bounds the jerk (e.g., restricted to $22.7\,\mathrm{m/s^3}$ in Scenario A) and maintains a smoother steering profile, thereby guaranteeing stable trackability during highly dynamic interactions.

\subsubsection{Algorithm-Level GP Efficiency}
\begin{figure}[t]
    \centering
    \includegraphics[width=\columnwidth]{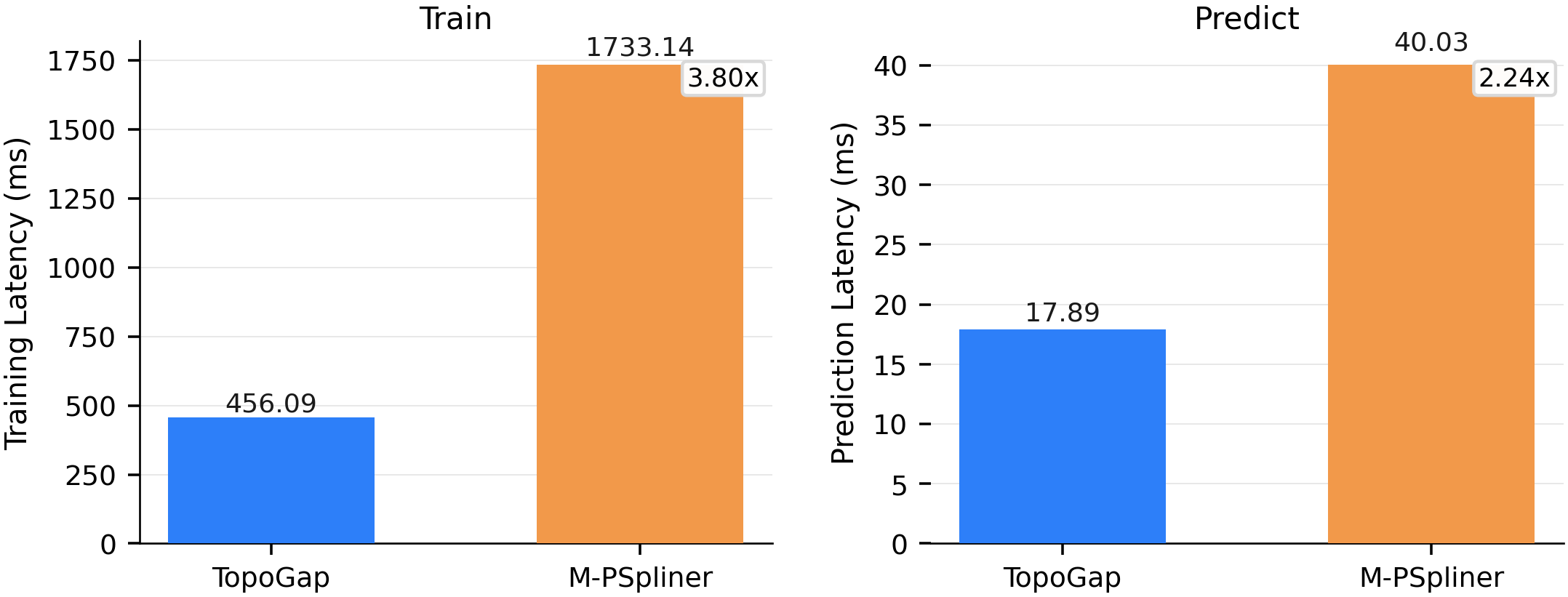}
    \caption{GP latency comparison: Topo-Gap achieves significant speedups over M-PSpliner in model training (left) and prediction (right).}
    \label{fig:final}
\end{figure}

 Evaluated under the standardized framework, Topo-Gap significantly outperforms M-PSpliner in execution speed (Fig. \ref{fig:final}). By leveraging the sparse design, our method yields 3.80$\times$ and 2.24$\times$  speedups in the GP training and prediction phases, respectively. This validates its fundamental efficiency advantage prior to full-system dynamic evaluations.
\subsubsection{Computational Efficiency and Solver Robustness}
\label{subsec:computation}
Based on the shadow mode protocol defined in Sec.~\ref{subsec:exp_setup}, the quantitative timing performance of our custom MPC solver against the baseline FSDP MPC solver is summarized in Table~\ref{tab:solver_compute}. Both solvers process the exact same dense spatial constraints, and our solver achieves the same level of optimization accuracy as the FSDP MPC solver in terms of objective cost.

However, their temporal distributions differ significantly. The baseline FSDP MPC solver suffers from structural update inefficiencies and heavy initialization overheads when handling highly constrained multi-agent corridors. This results in a pronounced long-tail latency, with its maximum MPC solve time spiking to $95.52\,\mathrm{ms}$. Such transient computational delays inevitably cause control starvation in the high-frequency autonomous racing loop. 

In contrast, our custom MPC solver is specifically designed to handle near-singular conditions and minimizes matrix setup overheads naturally. It reduces the average solve time to $19.48\,\mathrm{ms}$ and tightly bounds the worst-case maximum latency to $67.18\,\mathrm{ms}$. This deterministic temporal performance explicitly guarantees that the autonomous agent can continuously generate safe maneuvers without computational breakdown under extreme topological stress.

\begin{table}[htbp]
    \centering
    \caption{MPC Solve Time Benchmark (ms)}
    \label{tab:solver_compute}
    \resizebox{\columnwidth}{!}{%
    \begin{tabular}{l | c c c c}
    \toprule
    \textbf{Solver} & \textbf{Mean ($\mu_t$)} & \textbf{Std ($\sigma_t$)} & \textbf{99th Pct.$^*$} & \textbf{Max} \\
    \midrule
    FSDP MPC Solver & 24.44 & 14.46 & 75.96 & 95.52 \\
    \textbf{Our MPC Solver} & \textbf{19.48} & \textbf{10.90} & \textbf{53.90} & \textbf{67.18} \\
    \bottomrule
    \end{tabular}%
    }
    
    \vspace{0.3em}
    \parbox{\columnwidth}{\footnotesize $^*$99th Pct. denotes the 99th percentile of the solve time, indicating that 99\% of the optimizations are completed within this duration.}
    \vspace{-1.5em}
\end{table}
\section{Conclusion} \label{sec:conclusion}
In this paper, we propose Topo-Gap, a robust spatiotemporal motion planning framework tailored for the extreme dynamic conditions of unrestricted multi-agent autonomous racing. 
By generating opponents' SOC via parallel Sparse Gaussian Processes, our topology-aware gap identification proactively seizes overtaking windows while successfully suppressing decision oscillations. 
The trajectory is subsequently optimized by a custom PTC accelerated LTV-MPC to ensure strict kinematic feasibility.

Experimental results in the F1TENTH environment demonstrate the significant superiority of our approach. 
In baseline sequential overtaking scenarios, Topo-Gap reduces the total maneuver time by $51.6\%$ compared to SOTA predictive planners. 
Crucially, as topological complexity intensifies into dense platoons and tight bottlenecks, our framework consistently maintains a high overtaking success rate ($>81\%$) where baseline methods severely degrade or completely fail. 
Furthermore, our custom MPC solver reduces the average computational latency by $20.3\%$ and tightly bounds the worst-case delays, effectively eliminating control starvation at extreme racing speeds. 
Future work will explore the integration of game-theoretic interaction models to anticipate and safely navigate against active defensive maneuvers.


\bibliographystyle{IEEEtran}
\bibliography{reference}

\end{document}